\documentclass{article}
\usepackage{PRIMEarxiv}

\usepackage{url}
\usepackage[utf8]{inputenc} 
\usepackage[T1]{fontenc}    
\usepackage{hyperref}       
\usepackage{booktabs}       
\usepackage{amsfonts}       
\usepackage{nicefrac}       
\usepackage{microtype}      
\usepackage{xcolor}         
\usepackage{graphicx}
\usepackage{wrapfig}
\usepackage{amsmath}
\usepackage{mathtools}
\usepackage{multirow}
\usepackage{amssymb}
\usepackage{enumitem}
\usepackage{bm}
\usepackage[numbers,sort&compress]{natbib}
\usepackage{fancyhdr}
\title{Validation-Induced Shapley Shifts: How Validation Structure Distorts Data Valuation?}

\author{
  Yinan Shen \\
  Adobe \\
  Waltham, MA 02451, USA \\
  \texttt{yinans@adobe.com}
  \And
  Ziao Yang \\
  Department of Computer Science \\
  Brandeis University \\
  Waltham, MA 02453-2728, USA \\
  \texttt{ziaoyang@brandeis.edu}
  \And
  Hongfu Liu \\
  Department of Computer Science \\
  Brandeis University \\
  Waltham, MA 02453-2728, USA \\
  \texttt{hongfuliu@brandeis.edu}
}

\begin{document}


\maketitle

\begin{abstract}
Shapley values are widely used to attribute value to training data based on their marginal contribution to performance on a validation set. Existing practice often assumes these values are stable once the training data and model are fixed. In this work, we uncover a systematic vulnerability: even modest changes to the validation set—such as introducing noises—cause directional shifts in Shapley distributions. As noises are added, Shapley values of training samples compress toward zero. We trace this to a noise-induced neighborhood reshuffling effect: perturbations alter the local rank order between validation and training samples, flattening the valuation landscape. Using the KNN-Shapley framework, we show through synthetic and real data that these shifts are consistent and reproducible. Our findings challenge the assumption of Shapley stability and reveal a new axis of fragility in data valuation. We propose normalization and boundary-aware validation strategies to mitigate these distortions and enable more robust, interpretable valuation in machine learning marketplaces.
\end{abstract}

\section{Introduction}
Training data valuation is essential to collaborative learning, robust model training, and data marketplaces~\citep{sim2022data,ghorbani2019data,sim2020collaborative,agarwal2019marketplace}. Among various approaches, Shapley value–based methods provide a principled way to assign marginal credit to individual samples~\citep{shapley1953value,ghorbani2019data,jia2019efficient}, and have been widely used in pruning and noise detection~\citep{ghorbani2019data,schoch2022csshapley}, collaborative training~\citep{sim2020collaborative}, federated learning~\citep{wang2020principled,han2021data}, and active labeling~\citep{ghorbani2022active}. Their flexibility, axiomatic grounding, and model-agnostic nature make them a default choice for data valuation pipelines across domains.

While conceptually appealing, Shapley value computation is notoriously expensive. To address this, a line of work has focused on developing tractable variants for specific model classes, among which KNN-Shapley methods are especially attractive. These approaches leverage the structure of nearest-neighbor classifiers to compute exact or efficient Shapley values~\citep{jia2019efficient}, and more recent variants further improve interpretability and scalability by using soft-label formulations and linear-time variants~\citep{wang2023softknn, wangthreshold}. Other work explores correcting inflation bias~\citep{yang2024inflation} or optimizing pairwise interaction computation~\citep{belaid2023optimizing}, extending KNN-based Shapley beyond pointwise valuation. These advances make KNN-Shapley a practical and widely adopted testbed for data valuation analysis.

These methods rely on validation sets to compute utility, typically assuming they are fixed and clean. However, we show that structural changes—such as adding noise, or shifting boundary coverage—can systematically and directionally alter Shapley values, even under in-distribution perturbations. For example, injecting noise collapses values toward zero. These are not random effects, but consistent shifts driven by validation geometry—echoing recent concerns about inflation in KNN-Shapley~\citep{yang2024inflation}, data debugging~\citep{deutch2021explanations}, and distributional mismatch~\citep{ghorbani2020distributional}. This sensitivity raises concerns about the stability and fairness of Shapley-based valuations, especially in systems where validation design is decentralized or poorly standardized.

\looseness-1 This work does not introduce a new algorithm or model. Instead, we uncover an overlooked yet impactful phenomenon in data valuation and provide a geometric explanation for its cause. Our findings shed light on a critical source of distortion in Shapley-based attribution that has been largely ignored in prior work.
We summarize our contributions as follows:\vspace{2mm}
\begin{itemize}[wide=10pt, leftmargin=*, nosep]
\item We uncover a \textbf{phenomenon} that Shapley values are highly sensitive to in-distribution structural changes in the validation set, such as the injection of feature noise. These perturbations lead to consistent and directional shifts in the resulting valuation.\vspace{1mm}
\item We neighborhood reshuffling as the key \textbf{mechanism} driving these shifts. This mechanism links validation-set geometry to changes in local validation-to-training neighbor ranking, which in turn flattens and distorts marginal utility attribution.\vspace{1mm}
\item \looseness-1 We offer \textbf{actions} for fair data pricing and valuation design, including entropy-aware normalization and boundary-aware design. These insights improve the robustness and interpretability of Shapley-based systems in real-world marketplaces.
\end{itemize}

\section{Related Work}
\textbf{Data Valuation and Shapley Values}.
The Shapley value~\citep{shapley1953value} has become a widely used tool for quantifying the contribution of individual data points. Early work proposed scalable approximations~\citep{ghorbani2019data, jia2019efficient, jia2019towards}, with extensions to federated learning~\citep{wang2020principled}, and semi-supervised learning~\citep{courtnage2021shapleyssl}. Shapley-based methods have also been used in data pruning and selection~\citep{ghorbani2019data,schoch2022csshapley,ghorbani2022active,kwon2021beta}, feature importance~\citep{lundberg2017unified,covert2020improving}, and healthcare~\citep{pandl2021trustworthy,tang2021data}, where attribution accuracy is especially critical.
Recent advances include Beta Shapley~\citep{kwon2021beta}, Banzhaf value~\citep{wang2023data}, utility learning~\citep{wang2021improving}, and efficient computation and sampling~\citep{zhang2023efficient, castro2009polynomial, luo2022shapley}. Of special relevance is the KNN-Shapley line. \citet{jia2019efficient} showed closed-form computation is possible for KNN, inspiring further developments: threshold KNN~\citep{wangthreshold}, weighted KNN~\citep{wang2024efficient}, and soft-label variants~\citep{wang2023softknn}. 


\textbf{Robustness and Validation Set Design}.
Although grounded in marginal utility, Shapley values are practically sensitive to the validation set. Prior work has studied distributional and test-set mismatch~\citep{ghorbani2020distributional,recht2019imagenet}. Our work focuses on in-distribution structural perturbations—such as noise, and decision-boundary coverage—that can shift valuation outcomes even without dataset shift.

\textbf{Fairness in Machine Learning Markets}.
Incentive-compatible pricing and data markets have been explored through Shapley-based methods~\citep{agarwal2019marketplace, jia2019towards, ohrimenko2019collaborative}. These frameworks often assume fixed or unbiased utility. We challenge this assumption by showing that validation design can systematically distort attribution, highlighting fairness concerns in pricing and auditing when utility depends on factors like boundary density or sample entropy.

\section{Background: Data Valuation with Shapley Values}
Shapley values, rooted in cooperative game theory~\citep{shapley1953value}, provide a principled way to assign credit to individual data points based on their marginal contribution to model performance. In machine learning, they are widely used for tasks like pricing, selection, and auditing~\citep{ghorbani2019data, deutch2021explanations,agarwal2019marketplace}.

\textbf{Shapley Value for Training Data}.
Let $\mathcal{D}_{\text{train}}$ be a training dataset and $\mathcal{D}_{\text{val}}$ a validation set. The Shapley value of a training point $i$ is defined as:
\begin{equation}
\phi_i = \mathbb{E}_{S \subseteq \mathcal{D}_{\text{train}} \setminus \{i\}} \left[ U(S \cup \{i\}) - U(S) \right],
\end{equation}
where $U(S)$ measures model performance (e.g., average negative loss) on the validation set:
\begin{equation}
U(S) = - \frac{1}{|\mathcal{D}_{\text{val}}|} \sum\nolimits_{(x, y) \in \mathcal{D}_{\text{val}}} \mathcal{L}(f_S(x), y).
\end{equation}

\textbf{Efficient KNN-Shapley}.
For $k$-nearest neighbor models (KNN), \citet{jia2019efficient} showed that exact Shapley values can be computed efficiently. Building on this, \citet{wang2023softknn} proposed \emph{Soft-label KNN-SV}, which replaces hard labels with a soft utility formulation and supports efficient approximation via locality-sensitive hashing. This method improves robustness for small subsets and serves as the computational backbone of our study.

Although KNN-Shapley offers an efficient and interpretable framework, we uncover a surprising vulnerability: even small, in-distribution changes to the validation set—like adding noise—can cause significant and directional shifts in Shapley values. These are not random fluctuations, but structured effects driven by validation geometry. We investigate this phenomenon in depth in the next section.

\section{Phenomenon: Validation-Induced Shapley Shifts}
We begin our investigation with one synthetic dataset by injecting noises into the validation set, and uncover the phenomenon that these random changes on the validation structure lead to the directional shift towards the training samples' Shapley values. 

We generate a 2D balanced binary dataset using the \textit{make\_gaussian\_quantiles function} from sklearn.datasets, which samples from two overlapping Gaussians with class assignments based on quantiles. Feature noise is introduced by perturbing validation samples with additive Gaussian noise, simulating real-world data protection. Specifically, we perturb each validation point with Gaussian noise: $\epsilon \sim \mathcal{N}(0, \sigma^2 I)$, where $\sigma$ denotes the noise level.

\begin{figure}[h]
    \centering
    \includegraphics[width=0.6\linewidth]{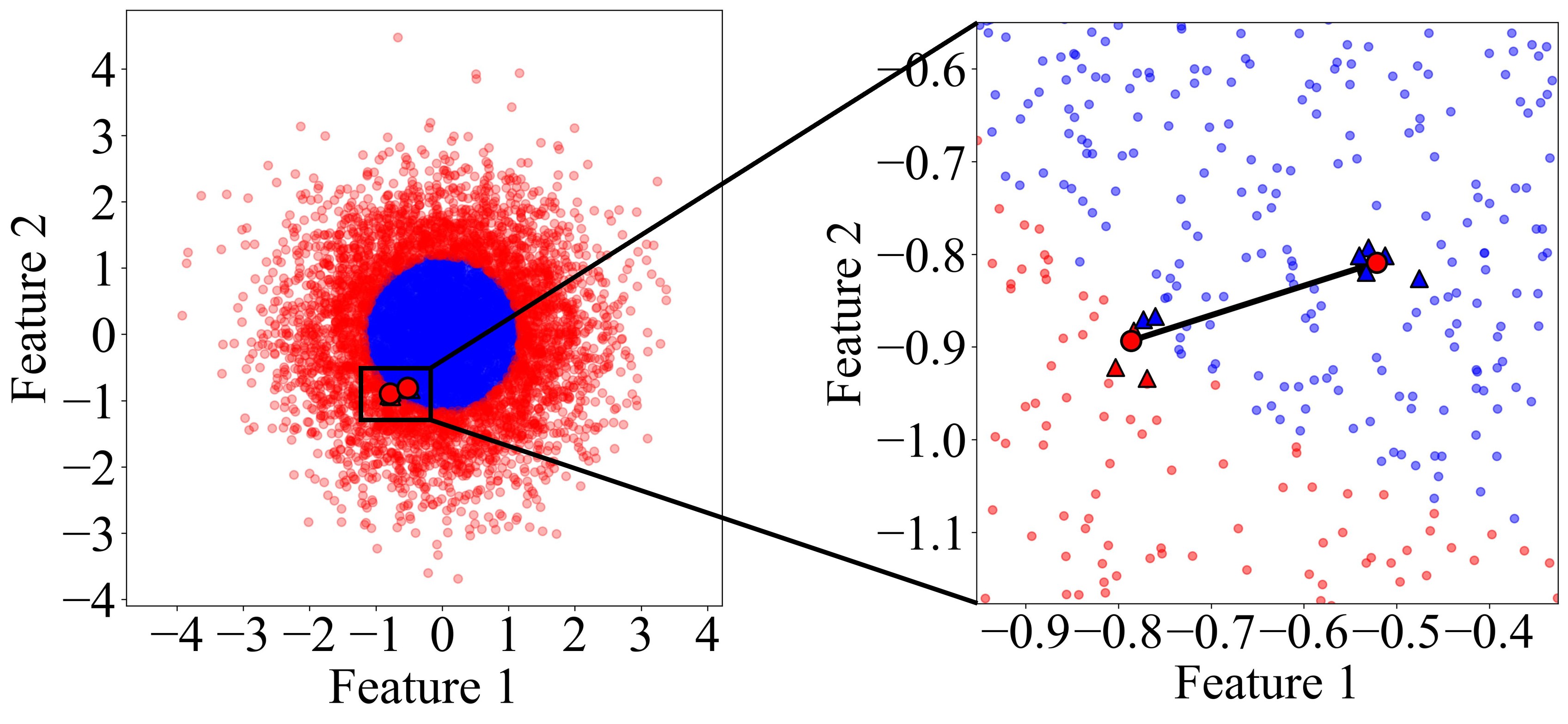}
    \caption{Noise injection demo. A validation sample near the decision boundary is perturbed by Gaussian noise. Colors indicate class labels. The perturbation here is small in magnitude but changes the sample's nearest neighbors, potentially altering its contribution signal.}
    \label{fig:noise}
\end{figure}

To illustrate the local impact of this perturbation, Figure~\ref{fig:noise} shows a synthetic 2D Gaussian dataset, where one validation sample near the class boundary is perturbed by Gaussian noise. Points are colored by ground-truth class. The left panel shows the distribution, while the right panel zooms into the neighborhood of the perturbed sample. This example highlights two key observations. First, although the perturbation is small, it is sufficient to change the sample’s nearest neighbors. Second, as a result, the perturbed sample may attribute utility to a different subset of training samples. This neighbor shift contributes to aggregate changes in Shapley values under noisy validation conditions.

\begin{figure}[t]
    \centering
    \includegraphics[width=0.8\linewidth]{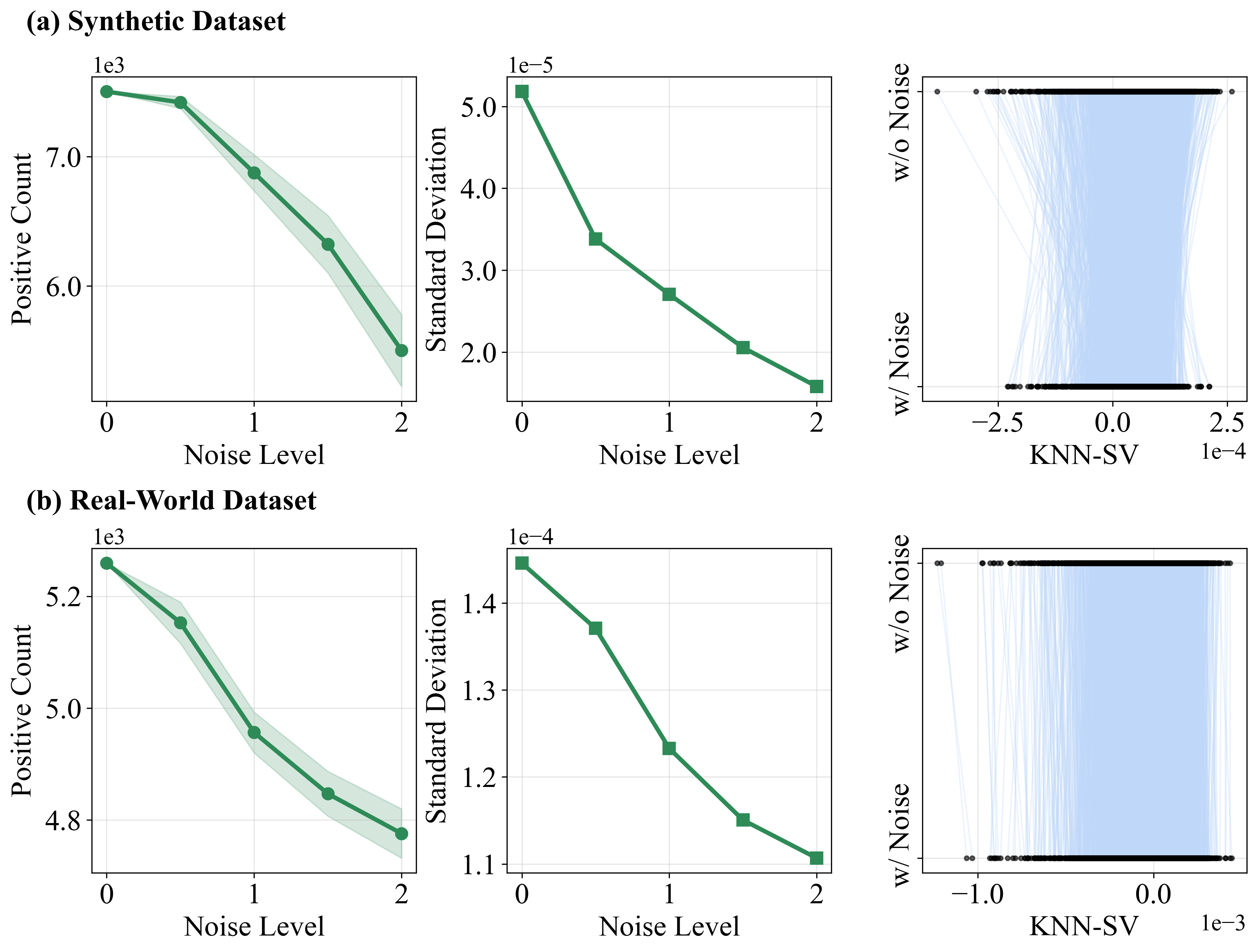}
    \caption{Shapley value shift under the noisy validation set. }
    \label{fig:shift}
\end{figure}

We now explore how injecting Gaussian noise into validation inputs—effectively creating slightly shifted versions of real samples—affects Shapley value distributions. As shown in Figure~\ref{fig:shift}, adding feature noise reduces the average Shapley value and further compresses variance, which leads the standard deviation of Shapley values becomes smaller. Moreover, the positive count of Shapley values becomes smaller with the increase of the noise level. Crucially, we observe the same directional shifts in real-world \textit{CreditCard} dataset~\citep{yeh2009creditcard}. Intuitively, these random noises lead unfixed effect; however, a clear pattern of the directional shift occur after the validation set is modified with noises. This can be also regarded as the cost of data protection.  

\section{Analysis: Mechanism via Boundary Samples}
To understand why Shapley values shift under the noisy validation set, we begin with fine-grained analyses at the level of individual training and validation samples. Since the Shapley value of a training sample is the average of its margin contribution on each validation sample, we first focus on the case with only one validation sample, and discover there are two distinct patterns according to whether the validation sample is located near a boundary. Further, we uncover how the noise on the valuation sample impact the training samples' Shapley values. 

\begin{figure}[t]
    \centering
    \includegraphics[width=0.8\linewidth]{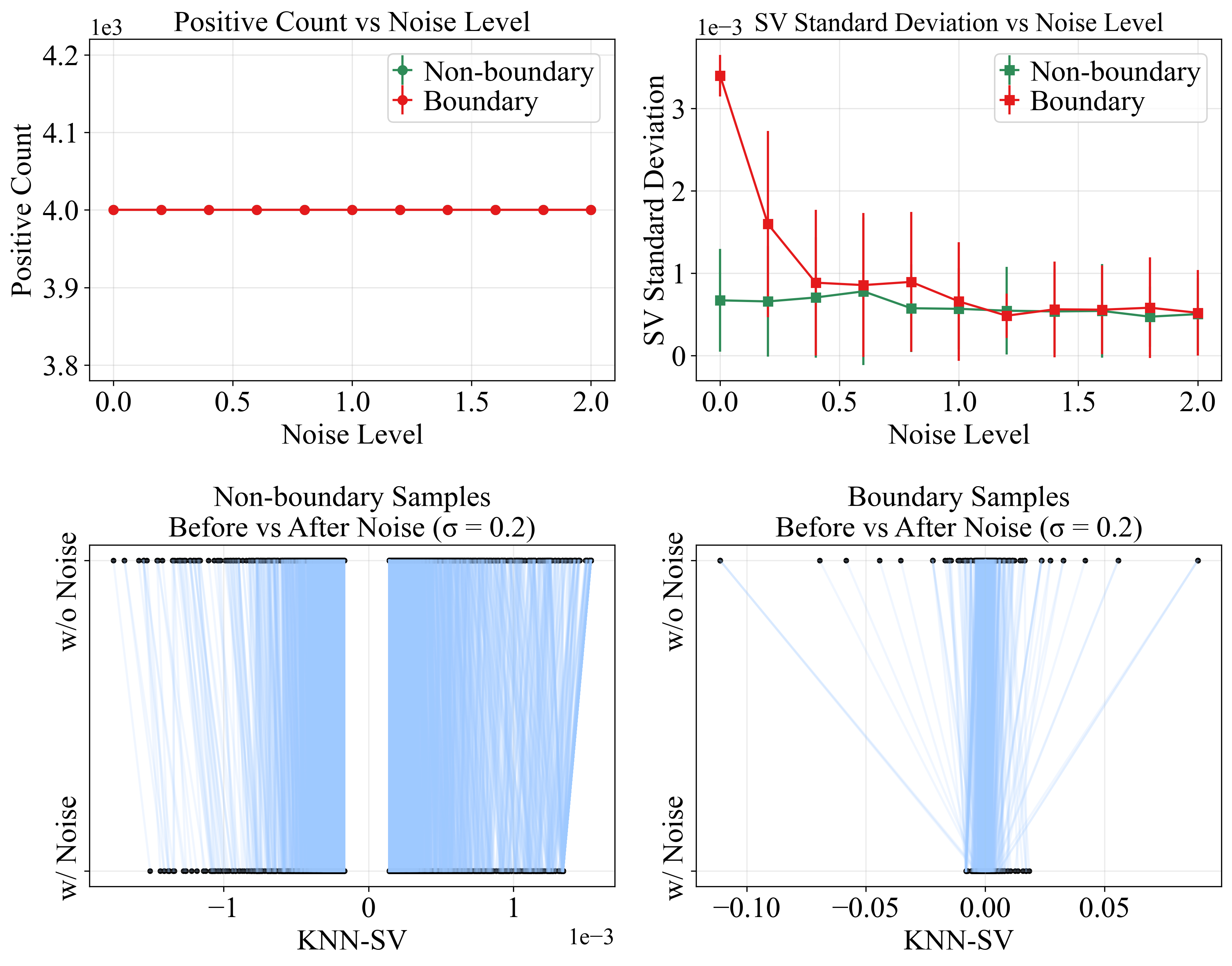}
    \caption{Case I with one validation sample. Boundary sample vs non-boundary sample. }
    \label{fig:one_sample}
\end{figure}

\textbf{Case I: One Validation Sample}. Figure~\ref{fig:one_sample} demonstrates two types of validation samples and the impact of noises on them. For a non-boundary validation sample, adding noises will not change the prediction; however, it might change the neighborhood structure, which we will discuss in next subsection. Therefore, we can see that the polarity of Shapley values of training samples keeps the same and the range of their values shrinks a little, but not too much. However, for a boundary validation sample, the noises make the original Shapley values of training samples shrink to the close-to-zero region, which contributes to the decrease of the standard deviation. 

We further explore the impact of different noise levels on positive counts and standard deviation. Since there is only one validation sample, there is no positive count change in this special case, no matter the validation sample is a boundary one or not. On the contrary, the standard deviation decreases significantly due to the boundary sample, while it keeps the still according to a non-boundary sample. This indicates that the shrink of the Shapley values accounts to the volatility of the boundary validation samples.

\textbf{Case II: Two Validation Samples}. We further analyze the cases with two validation samples of different categories. According to the boundary and non-boundary types, we consider three scenarios of two valuation samples, BB (two boundary validation samples), NN (two non-boundary validation samples), and BN (one boundary sample and one non-boundary sample).

\begin{figure}[t]
    \centering
    \includegraphics[width=0.8\linewidth]{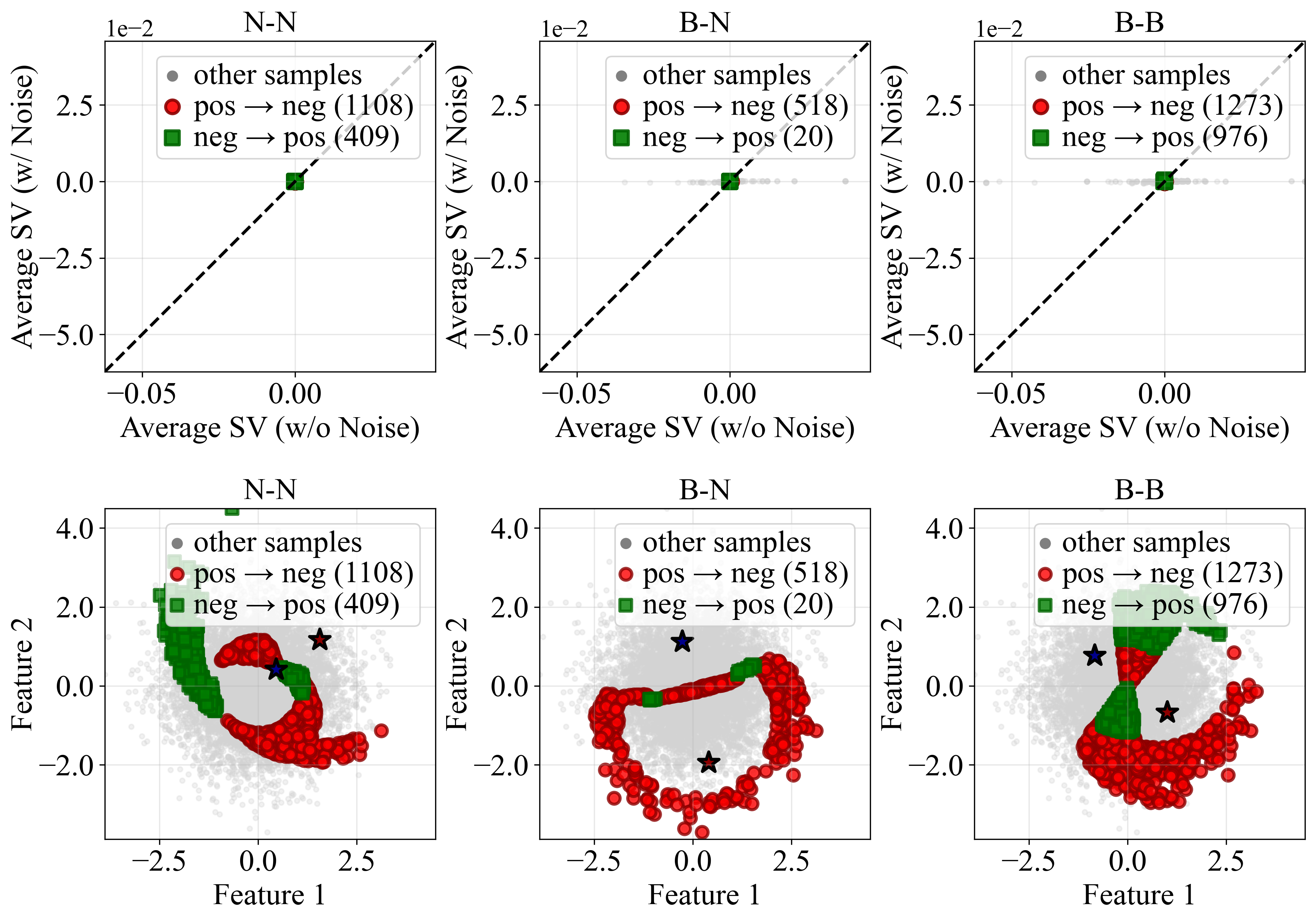}
    \caption{Case II with two validation samples. }
    \label{fig:two_samples}
\end{figure}

\begin{figure}[t]
    \centering
    \includegraphics[width=0.8\linewidth]{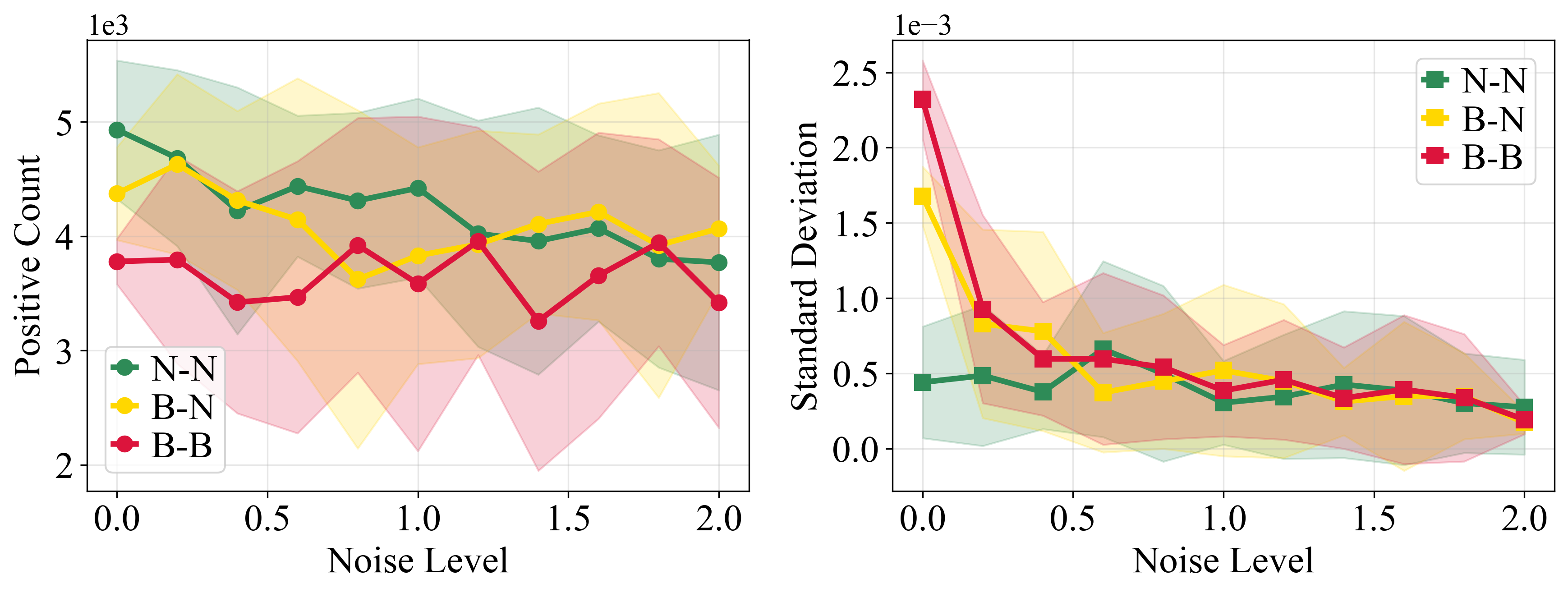}
    \caption{Case II with two validation samples. }
    \label{fig:two_samplesN}
\end{figure}

Figure~\ref{fig:two_samples} illustrates the three scenarios. The first row shows the change in Shapley values before and after noise is added, while the second row visualizes these changes in the original feature space. A key observation is that, across all three scenarios, there is a substantial decrease in positive Shapley values, which is much larger than the number of samples whose Shapley values switch from negative to positive after noise injection. Moreover, in the NN scenario, the reduction in positive Shapley values (around 700) is significantly larger than in the other two scenarios (approximately 500 and 300, respectively). This suggests that non-boundary samples are the primary factor driving the decrease in positive Shapley values. We further repeat the experiments under different noise ratios, and observe consistent trends, as shown in Figure~\ref{fig:two_samplesN}.

Combining the above two cases, we conclude that the observed phenomenon under noise injection in the validation set can be attributed to both boundary and non-boundary samples. Specifically, adding noise to boundary samples tends to narrow the range of Shapley values, while adding noise to non-boundary samples plays a dominant role in reducing positive Shapley values. Based on these, we will design actions for fair data pricing and valuation design.

\section{Action: Implications for Fair Data Pricing}
\looseness-1 To mitigate the sensitivity of Shapley values to noise in the validation set, we propose a principled correction framework for fair data valuation, including boundary-aware design. We organize the valuation result as a Shapley matrix $S \in \mathbb{R}^{n_{\rm train}\times n_{\rm val}}$, where row $i$ indexes a training sample and column $j$ indexes a validation sample. The validation columns are partitioned into a boundary group and a non-boundary group, and we compare two versions of this matrix: the clean baseline (no noise applied) and the noisy one. The full notation and correction details are deferred to Appendix~\ref{app:correction}.

Rather than correcting the full matrix at once, we first aggregate each training sample over the two validation groups, and then correct the two group scores separately. Let $s_i^{B,\mathrm{noisy}}$ and $s_i^{N,\mathrm{noisy}}$ denote the average contributions of training sample $i$ over the boundary and non-boundary groups under noise, respectively. We then rescale the two groups toward the baseline spread and apply a final bias correction:
\begin{align}
\bar s_i &= \lambda \Big(\mu_B^{\mathrm{noisy}} + \alpha_B\big(s_i^{B,\mathrm{noisy}} - \mu_B^{\mathrm{noisy}}\big)\Big) \\
&\quad + (1-\lambda)\Big(\mu_N^{\mathrm{noisy}} + \alpha_N\big(s_i^{N,\mathrm{noisy}} - \mu_N^{\mathrm{noisy}}\big)\Big),
\label{eq:action_main_mix}\\
\tilde s_i &= \bar s_i + b.
\label{eq:action_main_final}
\end{align}
Here $\lambda$ is the boundary fraction in the noisy validation set, $\mu_B^{\mathrm{noisy}}$ and $\mu_N^{\mathrm{noisy}}$ are the corresponding group means, $\alpha_B$ and $\alpha_N$ match the group-wise standard deviations to the baseline, and $b$ is chosen by quantile matching so that the fraction of positive values is restored to the baseline level. In this way, the boundary group mainly corrects the lost spread, while the non-boundary group mainly corrects the lost positive count.

\begin{figure}[t]
    \centering
    \includegraphics[width=0.8\linewidth]{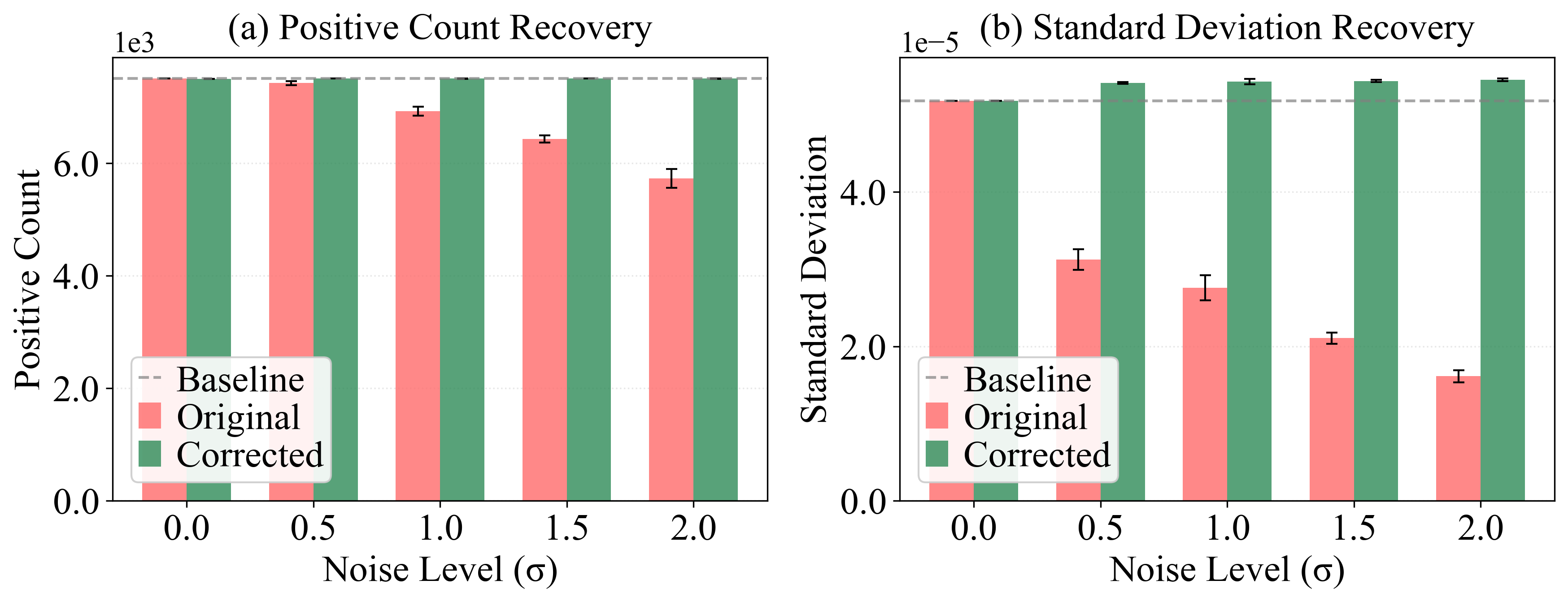}
    \caption{Shapley value correction on the Gaussian dataset. }
    \label{fig:action_s}
\end{figure}

\begin{figure}[t]
    \centering
    \includegraphics[width=0.8\linewidth]{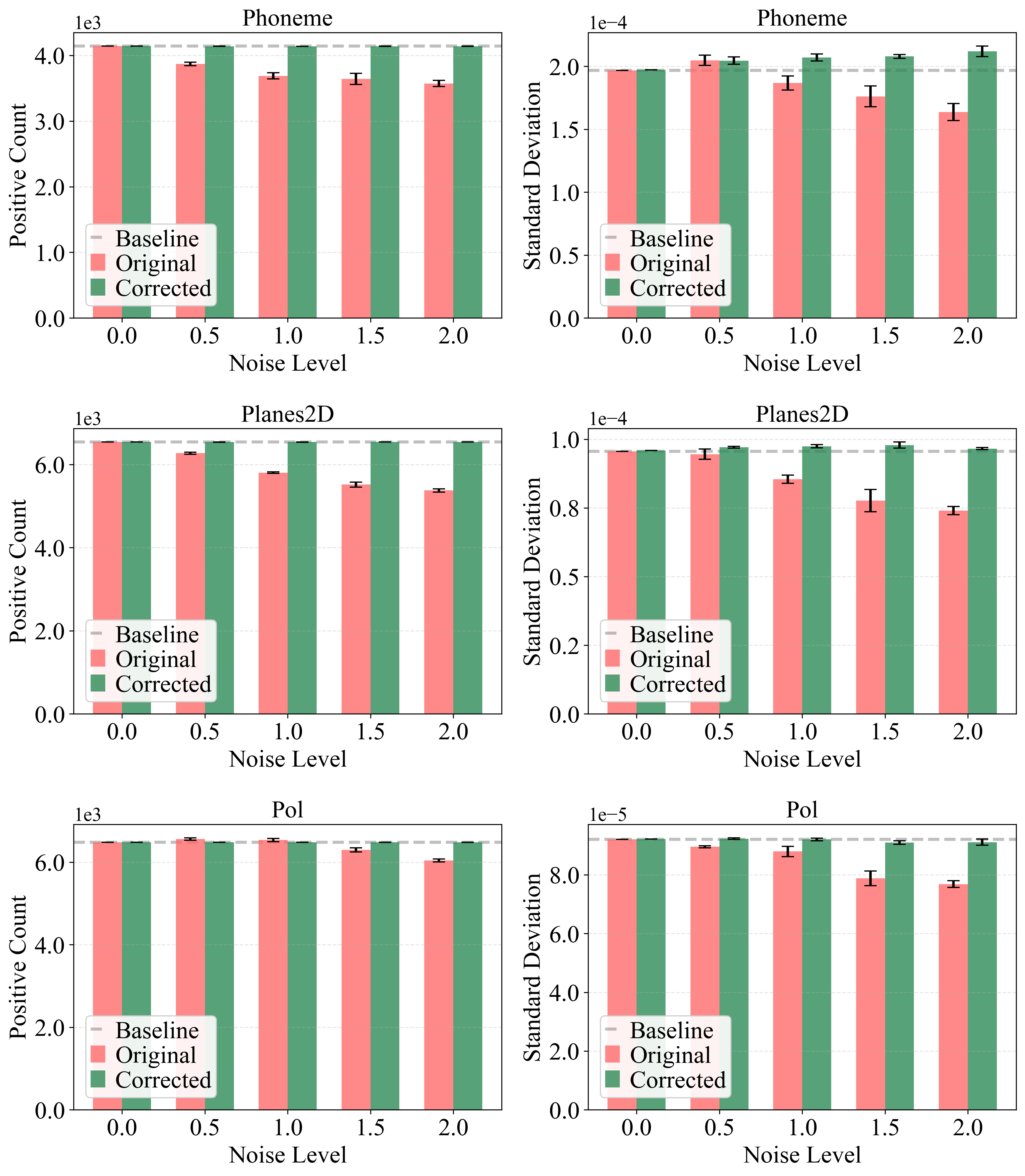}
    \caption{Shapley value correction on the real-world datasets: Phoneme, Planes2D, and Pol.}
    \label{fig:action_r1}
\end{figure}

\begin{figure}[t]
    \centering
    \includegraphics[width=0.8\linewidth]{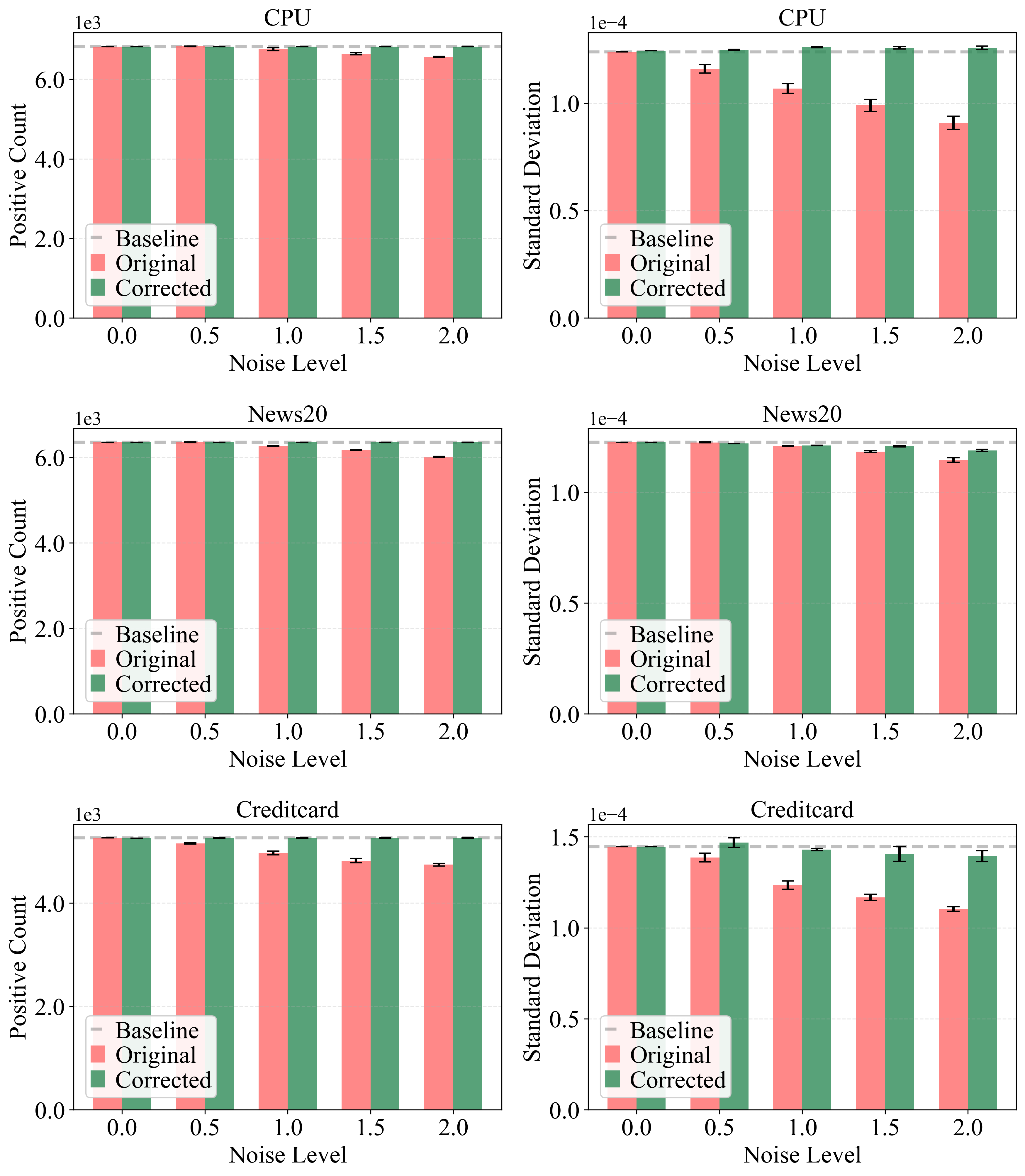}
    \caption{Shapley value correction on the real-world datasets: CPU, News20, and CreditCard.}
    \label{fig:action_r2}\vspace{-4mm}
\end{figure}

Figures~\ref{fig:action_s}--\ref{fig:action_r2} show that this correction substantially reduces the gap between noisy and baseline valuations on both the synthetic Gaussian data and six real-world datasets. Details of the real-world datasets are provided in Appendix~\ref{sec:realworld_datasets}. The corrected distribution tracks the baseline much more closely in both standard deviation and positive-count statistics, indicating that the boundary/non-boundary decomposition captures the main direction of the validation-induced shift. 

\section{Conclusion}
In this work, we revisit a fundamental assumption underlying Shapley-based data valuation: the stability of marginal contributions under a fixed model and training set. We show that this assumption breaks down in the presence of seemingly benign, in-distribution changes to the validation set. In particular, injecting noise induces consistent, directional shifts in Shapley values, leading to a systematic compression toward zero.

We attribute this behavior to an overlooked neighborhood reshuffling effect. By perturbing the local validation-to-training neighbor ranking, noise reduces the sharpness of contribution profiles. This effect manifests differently across validation regions: for boundary validation points, noise mainly induces positive and negative marginal contributions to cancel, whereas for non-boundary validation points, it makes positive contribution less concentrated and reduces the number of training points with net positive value. This mechanism provides a unified explanation for the observed distributional shifts and highlights the critical role of validation-set geometry in shaping data valuation.

\looseness-1 Our findings suggest that Shapley values are not purely intrinsic properties of training data, but are tightly coupled with the structure of the validation set used to evaluate utility. This insight has important implications for data-centric learning systems, particularly in decentralized or marketplace settings where validation protocols may vary or be manipulated.

\looseness-1 To address this fragility, we propose practical mitigation strategies with entropy-aware normalization and boundary-aware validation design to help restore stability and interpretability. More broadly, our work calls for a re-examination of evaluation protocols in data valuation, and opens up new directions for designing robust attribution methods.

\clearpage

\appendix
\section{Real-World Datasets}
\label{sec:realworld_datasets}

We evaluate the B/N correction method on six diverse real-world datasets, spanning speech recognition, spatial classification, political analysis, computer performance prediction, text classification, and financial fraud detection. Table~\ref{tab:datasets} summarizes their key characteristics. All experiments use KNN-Shapley with $K=5$ nearest neighbors, and noise is added to validation features only.

\textbf{Phoneme}: A binary classification task for distinguishing nasal from oral phonemes based on 5 acoustic features. The two classes correspond to whether the phoneme is produced with nasal or oral articulation~\citep{openmlphoneme1489}.

\textbf{Planes2D}: A spatial classification task from the OpenML benchmark. Each sample contains 10 continuous features describing spatial characteristics~\citep{breiman1984cart,openml2dplanes727}.

\textbf{Pol}: A political survey dataset containing 48 continuous features derived from political survey responses. The binary labels correspond to two political categories in the original survey data~\citep{openmlpol722}.

\textbf{CPU}: A computer activity prediction dataset with 21 continuous performance features. The task predicts whether the system is in a high or low activity state~\citep{delvecompactiv,openmlcpuact761}.

\textbf{News20}:
A text classification task where the original 20 newsgroups are reduced into a binary problem (computer-related vs. recreational topics). Each document is represented as a 384-dimensional Sentence-BERT embedding (\texttt{all-MiniLM-L6-v2})~\citep{lang1995newsweeder,reimers2019sentencebert}.

\textbf{Creditcard}: A credit card default prediction dataset with 23 numerical features including payment history, bill amounts, and demographic information. The binary labels distinguish defaulting versus non-defaulting clients~\citep{yeh2009creditcard}.

\begin{table}[h]
\centering
\caption{Real-world datasets used in our experiments.}
\label{tab:datasets}
\begin{tabular}{clll}
\toprule
\textbf{Dataset} & \textbf{Domain} & \textbf{Features} & \textbf{Labels} \\
\midrule
Phoneme   & Speech recognition    & 5   & nasal vs. oral \\
Planes2D  & Spatial classification & 10  & 2 classes \\
Pol       & Political analysis     & 48  & 2 classes \\
CPU       & System prediction      & 21  & high vs. low activity \\
News20    & Text classification    & 384 (SBERT) & computer vs. recreational \\
Creditcard& Credit default prediction        & 23  & default vs. non-default \\
\bottomrule
\end{tabular}
\end{table}

\section{Details of the Boundary-Aware Correction}
\label{app:correction}

\looseness-1 This appendix gives the full notation behind Eqs.~\eqref{eq:action_main_mix}--\eqref{eq:action_main_final}. We have two validation-set versions, $r\in\{\mathrm{bl},\mathrm{noisy}\}$, where "bl" denotes the baseline (no-noise) case and "noisy" refers to the case in which noise is applied to the validation samples. Let $S^r\in\mathbb{R}^{n_{\rm train}\times n_{\rm val}}$ denote the Shapley matrix, where $S^r_{ij}$ is the contribution of training sample $i$ to validation sample $j$. For each validation sample $j$, we inspect the labels of its $K=5$ nearest training neighbors and compute their label entropy. We use the resulting entropy rule to define a boundary mask $m^r\in\{0,1\}^{n_{\rm val}}$, where $m_j^r=1$ indicates that sample $j$ belongs to the boundary group and $m_j^r=0$ otherwise. Let $n_B^r=\lVert m^r\rVert_1$, $n_N^r=n_{\rm val}-n_B^r$, and $\lambda=n_B^{\mathrm{noisy}}/n_{\rm val}$.

For each training sample $i$, we aggregate its row of the Shapley matrix over the two groups using the mask element-wise:
\begin{equation}
 s_i^{B,r} = \frac{1}{n_B^r}\,\mathbf{1}^\top\!\big(S_{i:}^r \odot m^r\big),
 \qquad
 s_i^{N,r} = \frac{1}{n_N^r}\,\mathbf{1}^\top\!\big(S_{i:}^r \odot (\mathbf{1}-m^r)\big),
 \label{eq:action_group_scores}
\end{equation}
where $S_{i:}^r$ is the $i$-th row of $S^r$, $\odot$ denotes element-wise multiplication, and $\mathbf{1}$ is the all-one vector of length $n_{\rm val}$. We then summarize the distribution of these group scores across training samples by
\begin{equation}
\mu_G^r = \frac{1}{n_{\rm train}}\sum_{i=1}^{n_{\rm train}} s_i^{G,r},
\qquad
\sigma_G^r = \operatorname{Std}_{i}\!\big(s_i^{G,r}\big),
\qquad G\in\{B,N\}.
\label{eq:action_group_stats}
\end{equation}
The group-wise scaling factors are defined by
\begin{equation}
\alpha_G = \frac{\sigma_G^{\mathrm{bl}}}{\sigma_G^{\mathrm{noisy}}},
\qquad G\in\{B,N\}.
\label{eq:action_scale}
\end{equation}
Substituting Eq.~\eqref{eq:action_scale} into Eq.~\eqref{eq:action_main_mix} yields the bias-free corrected score $\bar s_i$.

To recover the sign structure of the clean baseline, we further match the positive ratio of the corrected distribution to that of the baseline. Let the overall baseline Shapley value of training sample $i$ be
\begin{equation}
 s_i^{\mathrm{bl}} = \frac{1}{n_{\rm val}}\,\mathbf{1}^\top S_{i:}^{\mathrm{bl}},
 \label{eq:action_baseline_sv}
\end{equation}
and define the baseline positive ratio as
\begin{equation}
\rho^{\mathrm{bl}} = \frac{1}{n_{\rm train}}\sum_{i=1}^{n_{\rm train}} \mathbb{I}\big[s_i^{\mathrm{bl}} > 0\big].
\label{eq:action_positive_ratio}
\end{equation}
Let $Q_q(\{\bar s_i\})$ denote the $q$-quantile of the set $\{\bar s_i\}_{i=1}^{n_{\rm train}}$. We set
\begin{equation}
 b = -Q_{1-\rho^{\mathrm{bl}}}(\{\bar s_i\}),
 \qquad
 \tilde s_i = \bar s_i + b.
 \label{eq:action_bias}
\end{equation}
This nearest-quantile matching ensures that the corrected scores recover the baseline fraction of positive Shapley values, while the group-wise scaling restores the spread contributed by the boundary and non-boundary groups.

\clearpage

\bibliography{ref}

\end{document}